# Bayesian Posterior Sampling via Stochastic Gradient Fisher Scoring

**Sungjin Ahn**                                                                 SUNGJIA@ICS.UCI.EDU
Dept. of Computer Science, UC Irvine, Irvine, CA 92697-3425, USA

**Anoop Korattikara**                                                           AKORATTI@ICS.UCI.EDU
Dept. of Computer Science, UC Irvine, Irvine, CA 92697-3425, USA

**Max Welling**                                                                 WELLING@ICS.UCI.EDU
Dept. of Computer Science, UC Irvine, Irvine, CA 92697-3425, USA

## Abstract

In this paper we address the following question: *"Can we approximately sample from a Bayesian posterior distribution if we are only allowed to touch a small mini-batch of data-items for every sample we generate?"*. An algorithm based on the Langevin equation with stochastic gradients (SGLD) was previously proposed to solve this, but its mixing rate was slow. By leveraging the Bayesian Central Limit Theorem, we extend the SGLD algorithm so that at high mixing rates it will sample from a normal approximation of the posterior, while for slow mixing rates it will mimic the behavior of SGLD with a pre-conditioner matrix. As a bonus, the proposed algorithm is reminiscent of Fisher scoring (with stochastic gradients) and as such an efficient optimizer during burn-in.

## 1. Motivation

When a dataset has a billion data-cases (as is not uncommon these days) MCMC algorithms will not even have generated a single (burn-in) sample when a clever learning algorithm based on stochastic gradients may already be making fairly good predictions. In fact, the intriguing results of Bottou and Bousquet (2008) seem to indicate that in terms of "number of bits learned per unit of computation", an algorithm as simple as stochastic gradient descent is almost optimally efficient. We therefore argue that for Bayesian methods to remain useful in an age when the datasets grow at an exponential rate, they need to embrace the ideas of the stochastic optimization literature.

A first attempt in this direction was proposed by Welling and Teh (2011) where the authors show that (uncorrected) Langevin dynamics with stochastic gradients (SGLD) will sample from the correct posterior distribution when the stepsizes are annealed to zero at a certain rate. While SGLD succeeds in (asymptotically) generating samples from the posterior at $O(n)$ computational cost with ($n \ll N$) it's mixing rate is unnecessarily slow. This can be traced back to its lack of a proper pre-conditioner: SGLD takes large steps in directions of small variance and reversely, small steps in directions of large variance which hinders convergence of the Markov chain. Our work builds on top of Welling and Teh (2011). We leverage the "Bayesian Central Limit Theorem" which states that when $N$ is large (and under certain conditions) the posterior will be well approximated by a normal distribution. Our algorithm is designed so that for large stepsizes (and thus at high mixing rates) it will sample from this approximate normal distribution, while at smaller stepsizes (and thus at slower mixing rates) it will generate samples from an increasingly accurate (non-Gaussian) approximation of the posterior. Our main claim is therefore that we can trade-in a usually small bias in our estimate of the posterior distribution against a potentially very large computational gain, which could in turn be used to draw more samples and reduce sampling variance.

From an optimization perspective one may view this algorithm as a Fisher scoring method based on stochastic gradients (see e.g. (Schraudolph et al., 2007)) but in such a way that the randomness introduced in the subsampling process is used to sample from the posterior distribution when we arrive at its mode. Hence, it is an efficient optimization algorithm that smoothly turns into a sampler when the correct (statistical) scale of precision is reached.

## 2. Preliminaries

We will start with some notation, definitions and preliminaries. We have a large dataset $X_N$ consisting of $N$ i.i.d. data-points $\{x_1...x_N\}$ and we use a family of distributions parametrized by $\theta \in \mathbb{R}^D$ to model the distribution of the $x_i$'s. We choose a prior distribution $p(\theta)$ and are interested in obtaining samples from the posterior distribution, $p(\theta|X_N) \propto p(X_N|\theta)p(\theta)$.

As is common in Bayesian asymptotic theory, we will also make use of some frequentist concepts in the development of our method. We assume that the true data generating distribution is in our family of models and denote the true parameter which generated the dataset $X_N$ by $\theta_0$. We denote the score or the gradient of the log likelihood w.r.t. data-point $x_i$ by $g_i(\theta) = g(\theta; x_i) = \nabla_\theta \log p(\theta; x_i)$. We denote the sum of scores of a batch of $n$ data-points $X_r = \{x_{r_1}...x_{r_n}\}$ by $G_n(\theta; X_r) = \sum_{i=1}^n g(\theta; x_{r_i})$ and the average by $\overline{g}_n(\theta; X_r) = \frac{1}{n} G_n(\theta; X_r)$. Sometimes we will drop the argument $X_r$ and instead simply write $G_n(\theta)$ and $\overline{g}_n(\theta)$ for convenience.

The covariance of the gradients is called the Fisher information defined as $I(\theta) = \mathbb{E}_x[g(\theta; x)g(\theta; x)^T]$, where $\mathbb{E}_x$ denotes expectation w.r.t the distribution $p(x; \theta)$ and we have used the fact that $\mathbb{E}_x[g(\theta; x)] = 0$. It can also be shown that $I(\theta) = -\mathbb{E}_x[H(\theta; x)]$, where H is the Hessian of the log likelihood.

Since we are dealing with a dataset with samples only from $p(x; \theta_0)$ we will henceforth be interested only in $I(\theta_0)$ which we will denote by $I_1$. It is easy to see that the Fisher information of $n$ data-points, $I_n = nI_1$. The empirical covariance of the scores computed from a batch of $n$ data-points is called the *empirical* Fisher information, $V(\theta; X_r) = \frac{1}{n-1} \sum_{i=1}^n (g_{r_i}(\theta) - \overline{g}_n(\theta))(g_{r_i}(\theta) - \overline{g}_n(\theta))^T$ (Scott, 2002). Also, it can be shown that $V(\theta_0)$ is a consistent estimator of $I_1 = I(\theta_0)$.

We now introduce an important result in Bayesian asymptotic theory. As $N$ becomes large, the posterior distribution becomes concentrated in a small neighbourhood around $\theta_0$ and becomes asymptotically Gaussian. This is formalized by the Bernstein-von Mises theorem, a.k.a the Bayesian Central Limit Theorem, (Le Cam, 1986), which states that under suitable regularity conditions, $p(\theta|\{x_1...x_N\})$ approximately equals $\mathcal{N}(\theta_0, I_N^{-1})$ as $N$ becomes very large.

## 3. Stochastic Gradient Fisher Scoring

We are now ready to derive our Stochastic Gradient Fisher Scoring (SGFS) algorithm. The starting point in the derivation of our method is the Stochastic Gradient Langevin Dynamics (SGLD) algorithm (Welling & Teh, 2011) which we describe in section 3.1. SGLD can sample accurately from the posterior but suffers from a low mixing rate. In section 3.2, we show that it is easy to construct a Markov chain that can sample from a normal approximation of the posterior at any mixing rate. We will then combine these methods to develop our Stochastic Gradient Fisher Scoring (SGFS) algorithm in section 3.3.

### 3.1. Stochastic Gradient Langevin Dynamics

The SGLD algorithm has the following update equation:

$$\theta_{t+1} \leftarrow \theta_t + \frac{\epsilon C}{2} \left\{ \nabla \log p(\theta_t) + N\overline{g}_n(\theta_t; X_n^t) \right\} + \nu$$
$$\text{where} \quad \nu \sim \mathcal{N}(0, \epsilon C) \quad (1)$$

Here $\epsilon$ is the step size, $C$ is called the preconditioning matrix (Girolami & Calderhead, 2010) and $\nu$ is a random variable representing injected Gaussian noise. The gradient of the log likelihood $G_N(\theta; X_N)$ over the whole dataset is approximated by scaling the mean gradient $\overline{g}_n(\theta_t; X_n^t)$ computed from a mini-batch $X_n^t = \{x_{t_1}...x_{t_n}\}$ of size $n \ll N$. Welling & Teh (2011) showed that Eqn. (1) generates samples from the posterior distribution if the step size is annealed to zero at a certain rate. As the step size goes to zero, the discretization error in the Langevin equation disappears and we do not need to conduct expensive Metropolis-Hasting(MH) accept/reject tests that use the whole dataset. Thus, this algorithm requires only $\mathcal{O}(n)$ computations to generate each sample, unlike traditional MCMC algorithms which require $\mathcal{O}(N)$ computations per sample.

However, since the step sizes are reduced to zero, the mixing rate is reduced as well, and a large number of iterations are required to obtain a good coverage of the parameter space. One way to make SGLD work at higher step sizes is to introduce MH accept/reject steps to correct for the higher discretization error, but our initial attempts using only a mini-batch instead of the whole dataset were unsuccessful.

### 3.2. Sampling from the Approximate Posterior

Since it is not clear how to use Eqn. (1) at high step sizes, we will move away from Langevin dynamics and explore a different approach. As mentioned in section 2, the posterior distribution can be shown to approach a normal distribution, $\mathcal{N}(\theta_0, I_N^{-1})$, as the size of the dataset becomes very large. It is easy to construct a Markov chain which will sample from this approximation of the posterior at any step size. We will now show that the following update equation achieves this:

$$\theta_{t+1} \leftarrow \theta_t + \frac{\epsilon C}{2} \left\{ -I_N(\theta_t - \theta_0) \right\} + \omega$$
$$\text{where} \quad \omega \sim \mathcal{N}(0, \epsilon C - \frac{\epsilon^2}{4} C I_N C) \quad (2)$$

The update is an affine transformation of $\theta_t$ plus injected independent Gaussian noise, $\omega$. Thus if $\theta_t$ has a Gaussian distribution $\mathcal{N}(\mu_t, \Sigma_t)$, $\theta_{t+1}$ will also have a Gaussian distribution, which we will denote as $\mathcal{N}(\mu_{t+1}, \Sigma_{t+1})$. These distributions are related by:

$$\mu_{t+1} = (I - \frac{\epsilon C}{2} I_N)\mu_t + \frac{\epsilon C}{2} I_N \theta_0$$
$$\Sigma_{t+1} = (I - \frac{\epsilon C}{2} I_N)\Sigma_t (I - \frac{\epsilon C}{2} I_N)^T + \epsilon C - \frac{\epsilon^2}{4} C I_N C \quad (3)$$

If we choose $C$ to be symmetric, it is easy to see that the approximate posterior distribution, $\mathcal{N}(\theta_0, I_N^{-1})$, is an invariant distribution of this Markov chain. Since Eqn. (2) is not a Langevin equation, it samples from the approximate posterior at large step-size and does not require any MH accept/reject steps. The only requirement is that $C$ should be symmetric and should be chosen so that the covariance matrix of the injected noise in Eqn. (2) is positive-definite.

### 3.3. Stochastic Gradient Fisher Scoring

In practical problems both sampling accuracy and mixing rate are important, and the extreme regimes dictated by both the above methods are very limiting. If the posterior is close to Gaussian (as is usually the case), we would like to take advantage of the high mixing rate. However, if we need to capture a highly non-Gaussian posterior, we should be able to trade-off mixing rate for sampling accuracy. One could also think about doing this in an "anytime" fashion where if the posterior is somewhat close to Gaussian, we can start by sampling from a Gaussian approximation at high mixing rates, but slow down the mixing rate to capture the non-Gaussian structure if more computation becomes available. In other words, one should have the freedom to manage the right trade off between sampling accuracy and mixing rate depending on the problem at hand.

With this goal in mind, we combine the above methods to develop our Stochastic Gradient Fisher Scoring (SGFS) algorithm. We accomplish this using a Markov chain with the following update equation:

$$\theta_{t+1} \leftarrow \theta_t + \frac{\epsilon C}{2} \left\{ \nabla \log p(\theta_t) + N \bar{g}_n(\theta_t; X_n^t) \right\} + \tau$$
$$\text{where} \quad \tau \sim \mathcal{N}(0, Q) \quad (4)$$

When the step size is small, we want to choose $Q = \epsilon C$ so that it behaves like the Markov chain in Eqn (1). Now we will see how to choose $Q$ so that when the step size is large and the posterior is approximately Gaussian, our algorithm behaves like the Markov chain in Eqn. (2). First, note that if $n$ is large enough for the central limit theorem to hold, we have:

$$\bar{g}_n(\theta_t; X_n^t) \sim \mathcal{N}\left(\mathbb{E}_x[g(\theta_t;x)], \frac{1}{n}\text{Cov}\left[g(\theta_t;x)\right]\right) \quad (5)$$

Here $\text{Cov}[g(\theta_t;x)]$ is the covariance of the scores at $\theta_t$. Using $N\text{Cov}[g(\theta_t;x)] \approx I_N$ and $N\mathbb{E}_x[g(\theta_t;x)] \approx G_N(\theta_t; X_N)$, we have:

$$\nabla \log p(\theta_t) + N\bar{g}_n(\theta_t; X_n^t)$$
$$\approx \nabla \log p(\theta_t) + G_N(\theta_t; X_N) + \phi$$
$$\text{where} \quad \phi \sim \mathcal{N}\left(0, \frac{NI_N}{n}\right) \quad (6)$$

Now, $\nabla \log p(\theta_t) + G_N(\theta_t; X_N) = \nabla \log p(\theta_t|X_N)$, the gradient of the log posterior. If we assume that the posterior is close to its Bernstein-von Mises approximation, we have $\nabla \log p(\theta_t|X_N) = -I_N(\theta_t - \theta_0)$. Using this in Eqn. (6) and then substituting in Eqn. (4), we have:

$$\theta_{t+1} \leftarrow \theta_t + \frac{\epsilon C}{2} \left\{ -I_N(\theta_t - \theta_0) \right\} + \psi + \tau \quad (7)$$

where,

$$\psi \sim \mathcal{N}\left(0, \frac{\epsilon^2}{4} \frac{N}{n} C I_N C\right) \quad \text{and} \quad \tau \sim \mathcal{N}(0, Q)$$

Comparing Eqn. (7) and Eqn. (2), we see that at high step sizes, we need:

$$Q + \frac{\epsilon^2}{4} \frac{N}{n} C I_N C = \epsilon C - \frac{\epsilon^2}{4} C I_N C \Rightarrow$$
$$Q = \epsilon C - \frac{\epsilon^2}{4} \frac{N+n}{n} C I_N C \quad (8)$$

Thus, we should choose Q such that:

$$Q = \begin{cases} \epsilon C & \text{for small } \epsilon \\ \epsilon C - \frac{\epsilon^2}{4}\gamma C I_N C & \text{for large } \epsilon \end{cases}$$

where we have defined $\gamma = \frac{N+n}{n}$. Since $\epsilon$ dominates $\epsilon^2$ when $\epsilon$ is small, we can choose $Q = \epsilon C - \frac{\epsilon^2}{4}\gamma C I_N C$ for both the cases above. With this, our update equation becomes:

$$\theta_{t+1} \leftarrow \theta_t + \frac{\epsilon C}{2} \left\{ \nabla \log p(\theta_t) + N\bar{g}_n(\theta_t; X_n^t) \right\} + \tau$$
$$\text{where} \quad \tau \sim \mathcal{N}\left(0, \epsilon C - \frac{\epsilon^2}{4}\gamma C I_N C\right) \quad (9)$$

Now, we have to choose $C$ so that the covariance matrix of the injected noise in Eqn. (9) is positive-definite. One way to enforce this, is by setting:

$$\epsilon C - \frac{\epsilon^2}{4}\gamma C I_N C = \epsilon C B C \Rightarrow C = 4\left[\epsilon \gamma I_N + 4B\right]^{-1} \quad (10)$$

where B is any symmetric positive-definite matrix. Plugging in this choice of $C$ in Eqn. 9, we get:

$$\theta_{t+1} \leftarrow \theta_t + 2\left[\gamma I_N + \frac{4B}{\epsilon}\right]^{-1} \times$$
$$\left\{ \nabla \log p(\theta_t) + N\bar{g}_n(\theta_t; X_t) + \eta \right\}$$
$$\text{where} \quad \eta \sim \mathcal{N}\left(0, \frac{4B}{\epsilon}\right) \quad (11)$$

However, the above method considers $I_N$ to be a known constant. In practice, we use $N\hat{I}_{1,t}$ as an estimate of $I_N$, where $\hat{I}_{1,t}$ is an online average of the empirical covariance of gradients (empirical Fisher information) computed at each $\theta_t$.

$$\hat{I}_{1,t} = (1 - \kappa_t)\hat{I}_{1,t-1} + \kappa_t V(\theta_t; X_n^t) \quad (12)$$

where $\kappa_t = 1/t$. In the supplementary material we prove that this online average converges to $I_1$ plus $\mathcal{O}(1/N)$ corrections if we assume that the samples are actually drawn from the posterior:

**Theorem 1.** *Consider a sampling algorithm which generates a sample $\theta_t$ from the posterior distribution of the model parameters $p(\theta|X_N)$ in each iteration $t$. In each iteration, we draw a random minibatch of size $n$, $X_n^t = \{x_{t_1}...x_{t_n}\}$, and compute the empirical covariance of the scores $V(\theta_t; X_n^t) = \frac{1}{n-1}\sum_{i=1}^{n}\{g(\theta_t; x_{t_i}) - \bar{g}_n(\theta_t)\}\{g(\theta_t; x_{t_i}) - \bar{g}_n(\theta_t)\}^T$. Let $V_T$ be the average of $V(\theta_t)$ across $T$ iterations. For large $N$, as $T \to \infty$, $V_T$ converges to the Fisher information $I(\theta_0)$ plus $\mathcal{O}(\frac{1}{N})$ corrections, i.e.*

$$\lim_{T \to \infty}\left[V_T \triangleq \frac{1}{T}\sum_{t=1}^{T}V(\theta_t; X_n^t)\right] = I(\theta_0) + \mathcal{O}(\frac{1}{N}) \quad (13)$$

Note that this is not a proof of convergence of the Markov chain to the correct distribution. Rather, assuming that the samples are from the posterior, it shows that the online average of the covariance of the gradients converges to the Fisher information (as desired). Thus, it strengthens our confidence that if the samples are almost from the posterior, the learned pre-conditioner converges to something sensible. What we do know is that if we anneal the stepsizes according to a certain polynomial schedule, and we keep the pre-conditioner fixed, then SGFS is a version of SGLD which was shown to converge to the correct equilibrium distribution (Welling & Teh, 2011). We believe the adaptation of the Fisher information through an online average is slow enough for the resulting Markov chain to still be valid, but a proof is currently lacking. The theory of adaptive MCMC (Andrieu & Thoms, 2009) or two time scale stochastic approximations (Borkar, 1997) might hold the key to such a proof which we leave for future work. Putting it all together, we arrive at algorithm 1 below.

The general method still has a free symmetric positive-definite matrix, $B$, which may be chosen according to our convenience. Examine the limit $\epsilon \to 0$. In this case our method becomes SGLD with preconditioning matrix $B^{-1}$ and step size $\epsilon$.

If the posterior is Gaussian, as is usually the case when $N$ is large, the proposed SGFS algorithm will sample correctly for arbitrary choice of $B$ even when the step size $\epsilon$ is large.

---

**Algorithm 1: Stochastic Gradient Fisher Scoring (SGFS)**

**Input:** $n$, $B$, $\{\kappa_t\}_{t=1:T}$
**Output:** $\{\theta_t\}_{t=1:T}$
1: Initialize $\theta_1$, $\hat{I}_{1,0}$
2: $\gamma \leftarrow \frac{n+N}{n}$
3: **for** $t = 1 : T$ **do**
4:    Choose random minibatch $X_n^t = \{x_{t_1}...x_{t_n}\}$
5:    $\bar{g}_n(\theta_t) \leftarrow \frac{1}{n}\sum_{i=1}^{n}g_{t_i}(\theta_t)$
6:    $V(\theta_t) \leftarrow \frac{1}{n-1}\sum_{i=1}^{n}\{g_{t_i}(\theta_t) - \bar{g}_n(\theta_t)\}\{g_{t_i}(\theta_t) - \bar{g}_n(\theta_t)\}^T$
7:    $\hat{I}_{1,t} \leftarrow (1 - \kappa_t)\hat{I}_{1,t-1} + \kappa_t V(\theta_t)$
8:    Draw $\eta \sim \mathcal{N}[0, \frac{4B}{\epsilon}]$
9:    $\theta_{t+1} \leftarrow \theta_t + 2\left(\gamma N\hat{I}_{1,t} + \frac{4B}{\epsilon}\right)^{-1}\{\nabla\log p(\theta_t) + N\bar{g}_n(\theta_t) + \eta\}$
10: **end for**

---

However, for some models the conditions of the Bernstein-von Mises theorem are violated and the posterior may not be well approximated by a Gaussian. This is the case for e.g. neural networks and discriminative RBMs, where the identifiability condition of the parameters do not hold. In this case, we have to choose a small $\epsilon$ to achieve accurate sampling (see section 5). These two extremes can be combined in a single "anytime" algorithm by slowly annealing the stepsize. For a non-adaptive version of our algorithm (i.e. where we would stop changing $\hat{I}_1$) after a fixed number of iterations) this would according to the results from Welling and Teh (2011) lead to a valid Markov chain for posterior sampling.

We recommend choosing $B \propto I_N$. With this choice, our method is highly reminiscent of "Fisher scoring" which is why we named it "Stochastic Gradient Fisher Scoring" (SGFS). In fact we can think of the proposed updates as a stochastic version of Fisher scoring based on small minibatches of gradients. But remarkably, the proposed algorithm is not only much faster than Fisher scoring (because it only requires small minibatches to compute an update), it also samples approximately from the posterior distribution. So the knife cuts on both sides: SGFS is a faster optimization algorithm but also doesn't overfit due to the fact that it switches to sampling when the right statistical scale of precision is reached.

## 4. Computational Efficiency

Clearly, the main computational benefit relative to standard MCMC algorithms comes from the fact that we use stochastic minibatches instead of the entire dataset at every iteration. However, for a model with a large number of parameters another source of significant computational effort is the computation of the $D \times D$ matrix $\gamma N\hat{I}_{1,t} + \frac{4B}{\epsilon}$ and multiplying its inverse with the mean gradient resulting in a total computational complexity of $\mathcal{O}(D^3)$ per iteration. In the case $n < D$ the computational complexity per iteration can be brought down to $\mathcal{O}(nD^2)$ by using the Sherman-Morrison-Woodbury equation. A more numerically stable alternative is to update Cholesky factors (Seeger, 2004).

In case even this is infeasible one can factor the Fisher information into $k$ independent blocks of variables of, say size $d$, in which case we have brought down the complexity to $\mathcal{O}(kd^3)$. The extreme case of this is when we treat every parameter as independent which boils down to replacing the Fisher information by a diagonal matrix with the variances of the individual parameters populating the diagonal. While for a large stepsize this algorithm will not sample from the correct Gaussian approximation, it will still sample correctly from the posterior for very small stepsizes. In fact, it is expected to do this more efficiently than SGLD which does not rescale its stepsizes at all. We have used the full covariance algorithm (SGFS-f) and the diagonal covariance algorithm (SGFS-d) in the experiments section.

## 5. Experiments

Below we report experimental results where we test SGFS-f, SGFS-d, SGLD, SGD and HMC on three different models: logistic regression, neural networks and discriminative RBMs. The experiments share the following practice in common. Stepsizes for SGD and SGLD are always selected through cross-validation for at least five settings. The minibatch size $n$ is set to either 300 or 500, but the results are not sensitive to the precise value as long as it is large enough for the central limit theorem to hold (typically, $n > 100$ is recommended). Also, we used $\kappa_t = \frac{1}{t}$.

### 5.1. Logistic Regression

A logistic regression model (LR) was trained on the MNIST dataset for binary classification of two digits 7 and 9 using a total of 10,000 data-items. We used a 50 dimensional random projection of the original features and ran SGFS with $\lambda = 1$. We used $B = \gamma I_N$ and tested the algorithm for a number of $\alpha$ values (where $\alpha = \frac{2}{\sqrt{\epsilon}}$). We ran the algorithm for 3,000 burn-in iterations and then collected 100,000 samples. We compare the algorithm to Hamiltonian Monte Carlo sampling (Neal, 1993) and to SGLD (Welling & Teh, 2011). For HMC, the "leapfrogstep" size was adapted during burn-in so that the acceptance ratio was around 0.8. For SGLD we also used a range of fixed stepsizes.

In figure 1 we show 2-d marginal distributions of SGFS compared to the ground truth from a long HMC run where we used $\alpha = 0$ for SGFS. From this we conclude that even for the largest possible stepsize the fit for SGFS-f is al-most perfect while SGFS-d underestimates the variance in this case (note however that for smaller stepsizes (larger $\alpha$) SGFS-d becomes very similar to SGLD and is thus guaranteed to sample correctly albeit with a low mixing rate).

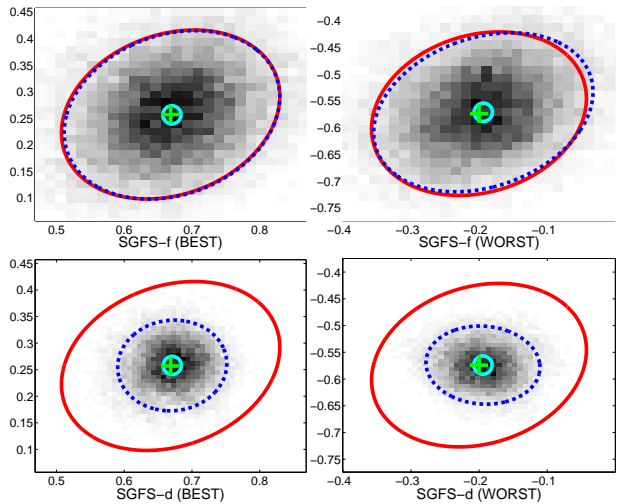

Figure 1. 2-d marginal posterior distributions for logistic regression. Grey colors correspond to samples from SGFS. Red solid and blue dotted ellipses represent iso-probability contours at two standard deviations away from the mean computed from HMC and SGFS, respectively. Top plots are the results for SGFS-f and bottom plots represent SGFS-d. Plots on the left represent the 2-d marginals with the smallest difference between HMC and SGFS while the plots on the right represent the 2-d marginals with the largest difference. Value for $\alpha$ is 0 meaning that no additional noise was added.

Next, we studied the inverse autocorrelation time per unit computation (ATUC)[1] averaged over the 51 parameters and compared this with the relative error after a fixed amount of computation time. The relative error is computed as follows: first we compute the mean and covariance of the parameter samples up to time $t$ : $\overline{\theta^t} = \frac{1}{t}\sum_{t'=1}^{t} \theta_{t'}$ and $C^t = \frac{1}{t}\sum_{t'=1}^{t}(\theta_{t'} - \overline{\theta^t})(\theta_{t'} - \overline{\theta^t})^T$. We do the same for the long HMC run which we indicate with $\overline{\theta^\infty}$ and $C^\infty$. Finally we compute

$$E_{1t} = \frac{\sum_i |\overline{\theta_i^t} - \overline{\theta_i^\infty}|}{\sum_i |\overline{\theta_i^\infty}|}, \quad E_{2t} = \frac{\sum_{ij} |C_{ij}^t - C_{ij}^\infty|}{\sum_{ij} |C_{ij}^\infty|} \quad (14)$$

In Figure 2 we plot the "Error at time T" for two values of T (T=100, T=3000) as a function of the *inverse* ATUC, which is a measure of the mixing rate. Top plots show the results for the mean and bottom plots for the covariance. Each point denoted by a cross is obtained from a different setting of parameters that control the mixing rate: $\alpha = [0, 1, 2, 3, 4, 5, 6]$ for SGFS, stepsizes $\epsilon = [1e-3, 5e-4, 1e-4, 5e-5, 1e-5, 5e-6, 1e-6]$ for SGLD, and number of leapfrog steps $s = [50, 40, 30, 20, 10, 1]$ for HMC. The circle is the result for the fastest mixing chain.

---

[1] ATUC = Autocorrelation Time × Time per Sample. Autocorrelation time is defined as $1 + 2\sum_{s=1}^{\infty} \rho(s)$ with $\rho(s)$ the autocorrelation at lag $s$ Neal (1993).

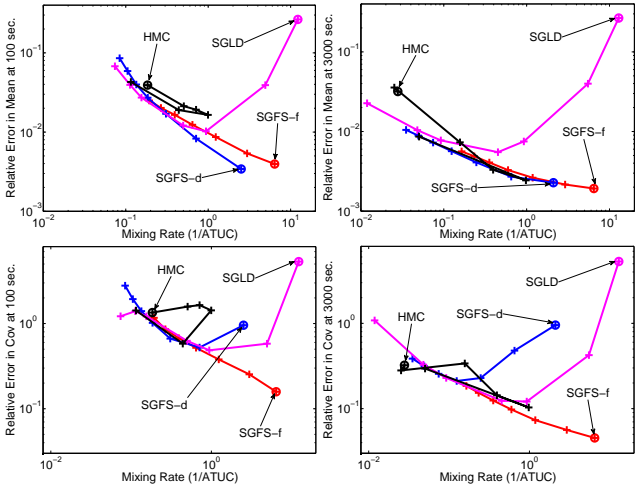

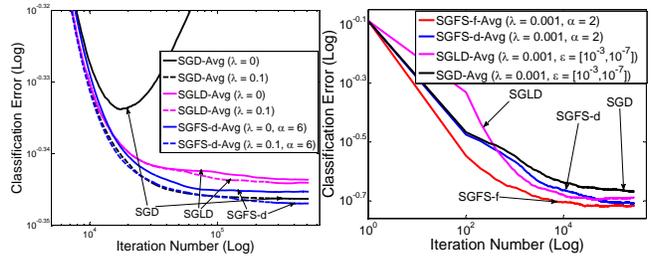

*Figure 3.* Test-set classification error of NNs trained with SGFS-f, SGFS-d, SGLD and SGD on the HHP dataset (left) and the MNIST dataset (right)

*Figure 2.* Final error of logistic regression at time T versus mixing rate for the mean (top) and covariance (bottom) estimates after 100 (left) and 3000 (right) seconds of computation. See main text for detailed explanation.

For SGFS and SGLD, if the slope of the curve is negative (downward trend) then the corresponding algorithm was still in the phase of reducing error by reducing sampling variance at time T. However, when the curve bends upwards and develops a positive slope the algorithm has reached its error floor corresponding to the approximation bias. The situation is different for HMC, (which has no bias) but where the bending occurs because the number of leapfrog steps has become so large that it is turning back on itself. HMC is not faring well because it is computationally expensive to run (which hurts both its mixing rate and error at time T). We also observe that in the allowed running time SGFS-f has not reached its error floor (both for the mean and the covariance). SGFS-d is reaching its error floor only for the covariance (which is consistent with Figure 1 bottom) but still fares well in terms of the mean. Finally, for SGLD we clearly see that in order to obtain a high mixing rate (low ATUC) it has to pay the price of a large bias. These plots clearly illustrate the advantage of SGFS over both HMC as well as SGLD.

**5.2. SGFS on Neural Networks**

We also applied our methods to a 3 layer neural network (NN) with logistic activation functions. Below we describe classification results for two datasets.

### 5.2.1. HERITAGE HEALTH PRIZE (HHP)

The goal of this competition is to predict how many days between $[0 - 15]$ a person will stay in a hospital given his/her past three years of hospitalization records[2]. We used the same features as the team *market makers* that won the first milestone prize. Integrating the first and second year data, we obtained 147,473 data-items with 139 feature dimensions and then used a randomly selected 70% for training and the remainder for testing. NNs with 30 hidden units were used because more hidden units did not noticeably improve the results. Although we used $\alpha = 6$ for SGFS-d, there was no significant difference for values in the range $3 \leq \alpha \leq 6$. However, $\alpha < 3$ did not work for this dataset due to the fact that many features had values 0.

For SGD, we used stepsizes from a polynomial annealing schedule $a(b + t)^{-\delta}$. Because the training error decreased slowly in a valid range $\delta = [0.5, 1]$, we used $\delta = 3$, $a = 10^{14}$, $b = 2.2 \times 10^5$ instead which was found optimal through cross-validation. (This setting reduced the stepsize from $10^{-2}$ to $10^{-6}$ during 1e+7 iterations). For SGLD, $a = 1$, $b = 10^4$, and $\delta = 1$ reducing the step size from $10^{-4}$ to $10^{-6}$ was used. Figure 3 (left) shows the classification errors averaged over the posterior samples for two regularizer values, $\lambda = 0$ and the best regularizer value $\lambda$ found through cross-validation. First, we clearly see that SGD severely overfits without a regularizer while SGLD and SGFS prevent it because they average predictions over samples from a posterior mode. Furthermore, we see that when the best regularizer is used, SGFS (marginally) outperforms both SGD and SGLD. The result from SGFS-d submitted to the actual competition leaderboard gave us an error of 0.4635 which is comparable to 0.4632 obtained by the milestone winner with a fine-tuned Gradient Boosting Machine.

---

[2] http://www.heritagehealthprize.com

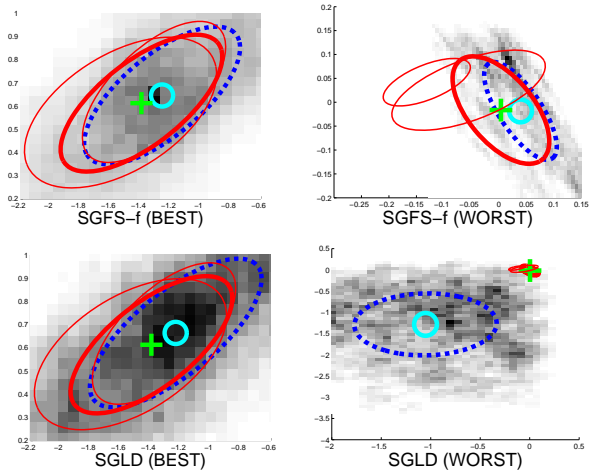

Figure 4. 2-d marginal posterior distributions of DRBM. Grey colors correspond to samples from SGFS/SGLD. Thick red solid lines correspond to iso-probability contours at two standard deviations away from the mean computed from HMC samples. Thin red solid lines correspond to HMC results based on subsets of the samples. The thick blue dashed lines correspond to SGFS-f (top) and SGLD (bottom) runs. Plots on the left represent the 2-d marginals with the smallest difference between HMC and SGFS/SGLD while the plots on the right represent the 2-d marginals with the largest difference.

### 5.2.2. CHARACTER RECOGNITION

We also tested our methods on the MNIST dataset for 10 digit classification which has 60,000 training instances and 10,000 test instances. In order to test with SGFS-f, we used inputs from 20 dimensional random projections and 30 hidden units so that the number of parameters equals 940. Moreover, we increased the mini-batch size to 2,000 to reduce the time required to reach a good approximation of the $940 \times 940$ covariance matrix. The classification error averaged over the samples is shown in Figure 3 (right). Here, we used a small regularization parameter of $\lambda = 0.001$ for all methods as overfitting was not an issue. For SGFS, $\alpha = 2$ is used while for both SGD and SGLD the stepsizes were annealed from $10^{-3}$ to $10^{-7}$ using $a = 1$, $b = 1000$, and $\gamma = 1$.

### 5.3. Discriminative Restricted Boltzmann Machine (DRBM)

We trained a DRBM (Larochelle & Bengio, 2008) on the KDD99 dataset which consists of 4,898,430 datapoints with 40 features, belonging to a total of 23 classes. We first tested the classification performance by training the DRBM using SGLD, SGFS-f, SGFS-d and SGD. For this experiment the dataset was divided into a 90% training set, 5% validation and 5% test set. We used 41 hidden units giving us a total of 2647 parameters in the model. We used $\lambda = 10$ and $B = \gamma I_N$. We tried 6 different $(\alpha, \epsilon)$ combinations for SGFS-f and SGFS-d and tried 18 annealing schedules for SGD and SGLD, and used the validation set to pick the best one. The best results were obtained with an $\alpha$ value of 8.95 for SGFS-f and SGFS-d, and [$a = 0.1$, $b = 100000$, $\delta = 0.9$] for SGD and SGLD. We ran all algorithms for 100,000 iterations. Although we experimented with different burn-in iterations, the algorithms were insensitive to this choice. The final error rates are given in table 1 from which we conclude that the samplers based on stochastic gradients can act as effective optimizers whereas HMC on the full dataset becomes completely impractical because it has to compute 11.7 billion gradients per iteration which takes around 7.5 minutes per sample (4408587 datapoints × 2647 parameters).

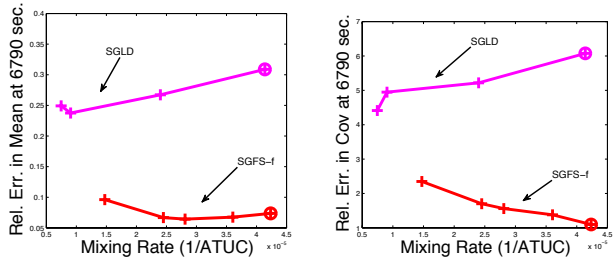

Figure 5. Final error for DRBM at time T versus mixing rate for the mean (left) and covariance (right) estimates after 6790 seconds of computation on a subset of KDD99.

| SGD | SGLD | SGFS-d | SGFS-f |
|---|---|---|---|
| $8.010^{-4}$ | $6.610^{-4}$ | $4.210^{-4}$ | $4.410^{-4}$ |

Table 1. Final test error rate on the KDD99 dataset.

To compare the quality of the samples drawn after burn-in, we created a 10% subset of the original dataset. This time we picked only the 6 most populous classes. We tested all algorithms with 41, 10 and 5 hidden units, but since the posterior is highly multi-modal, the different algorithms ended up sampling from different modes. In an attempt to get a meaningful comparison, we therefore reduced the number of hidden units to 2. This improved the situation to some degree, but did not entirely get rid of the multi-modal and non-Gaussian structure of the posterior. We compare results of SGFS-f/SGLD with 30 independent HMC runs, each providing 4000 samples for a total of 120,000 samples. Since HMC was very slow (even on the reduced set) we initialized at a mode and used the Fisher information at the mode as a pre-conditioner. We used 1 leapfrog step and tuned the step-size to get an acceptance rate of 0.8. We ran SGFS-f with $\alpha = [2, 3, 4, 5, 10]$ and SGLD with fixed step sizes of [5e-4, 1e-4, 5e-5, 1e-5, 5e-6]. Both algorithms were initialized at the same mode and ran for 1 million iterations. We looked at the marginal distribu-tions of the top 25 pairs of variables which had the highest correlation coefficient. In Figure 4 (top-left and bottom-left) we show a set of parameters where both SGFS-f and SGLD obtained an accurate estimate of the marginal posterior. In 4 (top-right and bottom-right) we show an example where SGLD failed. The thin solid red lines correspond to HMC runs computed from various subsets of the samples, whereas the thick solid red line is computed using the all samples from all HMC runs. We have shown marginal posterior estimates of the SGFS-f/SGLD algorithms with a thick dashed blue ellipse. After inspection, it seemed that the posterior structure was highly non-Gaussian with regions where the probability very sharply decreased. SGLD regularly stepped into these regions and then got catapulted away due to the large gradients there. SGFS-f presumably avoided those regions by adapting to the local covariance structure. We found that in this region even the HMC runs are not consistent with one another. Note that the SGFS-f contours seem to agree with the HMC contours as much as the HMC contours agree with the results of its own subsets, in both the easy and the hard case.

Finally, we plot the error after 6790 seconds of computation versus the mixing rate. Figure 5-left shows the results for the mean and the right for the covariance (for an explanation of the various quantities see discussion in section 5.1). We note again that SGLD incurs a significantly larger approximation bias at the same mixing rate as SGFS-f.

## 6. Conclusions

We have introduced a novel method, "Stochastic Gradient Fisher Scoring" (SGFS) for approximate Bayesian learning. The main idea is to use stochastic gradients in the Langevin equation and leverage the central limit theorem to estimate the noise induced by the subsampling process. This subsampling noise is combined with artificially injected noise and multiplied by the estimated inverse Fisher information matrix to approximately sample from the posterior. This leads to the following desirable properties.

• Unlike regular MCMC methods, SGFS is fast because it uses only stochastic gradients based on small mini-batches to draw samples.
• Unlike stochastic gradient descent, SGFS samples (approximately) from the posterior distribution.
• Unlike SGLD, SGFS samples from a Gaussian approximation of the posterior distribution (that is correct for $N \to \infty$) for large stepsizes.
• By annealing the stepsize, SGFS becomes an anytime method capturing more non-Gaussian structure with smaller stepsizes but at the cost of slower mixing.
• During its burn-in phase, SGFS is an efficient optimizer because like Fisher scoring and Gauss-Newton methods, it is based on the natural gradient.

For an appropriate annealing schedule, SGFS thus goes through three distinct phases: 1) during burn-in we use a large stepsize and the method is similar to a stochastic gradient version of Fisher scoring, 2) when the stepsize is still large, but when we have reached the mode of the distribution, SGFS samples from the asymptotic Gaussian approximation of the posterior, and 3) when the stepsize is further annealed, SGFS will behave like SGLD with a pre-conditioning matrix and generate increasingly accurate samples from the true posterior.

## Acknowledgements

This material is based upon work supported by the National Science Foundation under Grant No. 0447903, 0914783, 0928427.